\pdfoutput=1
\documentclass[journal]{IEEEtran}
\usepackage{multirow}
\usepackage{booktabs}
\usepackage{graphicx}
\usepackage[utf8]{inputenc}  
\usepackage[T1]{fontenc}     
\usepackage{amsmath}
\usepackage{url}
\usepackage{soul}
\usepackage{siunitx}
\usepackage[useregional,showdow=false]{datetime2}
\usepackage[colorlinks=true, linkcolor=black, citecolor=black, urlcolor=blue]{hyperref} 

\ifCLASSINFOpdf
\else
\fi

\usepackage{float}
\begin{document}
\title{Multispectral LiDAR data for extracting tree points in urban and suburban areas}

\author{Narges~Takhtkeshha,
        Gabriele~Mazzacca,
        Fabio~Remondino,
        Juha~Hyyppä,
        and Gottfried~Mandlburger

\thanks{Narges Takhtkeshha and Gottfried Mandlburger are with the Department of Geodesy and Geoinformation, TU Wien, Vienna 1040, Austria (e-mail: Narges.Takhtkeshha@geo.tuwien.ac.at; gottfried.mandlburger@geo.tuwien.ac.at)}
\thanks{Narges Takhtkeshha, Gabriele Mazzacca, and Fabio Remondino are with 3D Optical Metrology (3DOM) unit, Bruno Kessler Foundation (FBK), Trento 38123, Italy (e-mail:ntakhtkeshha@fbk.eu; gmazzacca@fbk.eu; remondino@fbk.eu)}
\thanks{Juha Hyyppä is with Department of Photogrammetry and Remote Sensing, Finnish Geospatial Research Institute, National Land Survey of Finland, Espoo 02150, Finland (e-mail:juha.hyyppa@nls.fi)}

\thanks{Manuscript received \today; revised \today}}

\maketitle
\begin{abstract}Monitoring urban tree dynamics is vital for supporting greening policies and reducing risks to electrical infrastructure. Airborne laser scanning has advanced large-scale tree management, but challenges remain due to complex urban environments and tree variability. Multispectral (MS) light detection and ranging (LiDAR) improves this by capturing both 3D spatial and spectral data, enabling detailed mapping. This study explores tree point extraction using MS-LiDAR and deep learning (DL) models. Three state-of-the-art models are evaluated: Superpoint Transformer (SPT), Point Transformer V3 (PTv3), and Point Transformer V1 (PTv1). Results show the notable time efficiency and accuracy of SPT, with a mean intersection over union (mIoU) of 85.28\%. The highest detection accuracy is achieved by incorporating pseudo normalized difference vegetation index (pNDVI) with spatial data, reducing error rate by 10.61 percentage points (pp) compared to using spatial information alone. These findings highlight the potential of MS-LiDAR and DL to improve tree extraction and further inventories.

\end{abstract}

\IEEEpeerreviewmaketitle
\section{Introduction}
\IEEEPARstart{R}{apid} urbanization has led to a range of environmental and urban challenges, especially urban heat island effects and air pollution. Trees are a vital component of urban green infrastructure, being essential for the health and well-being of urban residents. Therefore, their monitoring is crucial. Thanks to the exceptional penetration capabilities and consequently rich 3D structural information captured by LiDAR sensors, point cloud data have attracted increasing attention for monitoring trees in both urban and forest environments \cite{yang2024}. An essential preliminary step in urban tree inventory is the removal of non-tree points from the point cloud. However, the complexity and heterogeneity of urban scenes make this task particularly challenging.

In recent years, driven by evolving application demands and the need for higher accuracy, LiDAR technology has progressed from single-channel systems to multispectral and hyperspectral systems \cite{Hyypaa2013, takhtkeshha2024}. These advanced systems are capable of simultaneously capturing 3D spatial and spectral information, offering a single data source solution with promising results, especially for tree inventory \cite{hakula2023,RUOPPA2025}.

 Tree point extraction methods can generally be classified into two categories: raster-based and point-based. Unlike raster-based approaches, point-based methods avoid the occlusion issues inherent in 2D images and provide more precise differentiation between low and medium/high vegetation — an essential factor for accurate tree inventory applications. Various heuristic, machine, and deep learning (DL) algorithms have been used for point-based tree extraction, with DL methods demonstrating superior performance \cite{adityaBenchmark2024,yang2024}. Generally, class-specific DL models are increasingly receiving attention due to the diversity of objects across different locations and class definition inconsistencies, while also requiring less detailed data annotation \cite{adityaGreenSegNet2024}. The shadow-free spatial-spectral information provided by MS-LiDAR data holds promise for enhancing tree point extraction. Accordingly, the primary motivation of this work is to leverage recent state-of-the-art transformer-based deep learning models and MS-LiDAR data to effectively separate tree and non-tree points in complex urban environments through binary semantic segmentation, thereby improving the accuracy of subsequent individual tree segmentation and inventory.

 

\subsection{Related Work}
MS-LiDAR has demonstrated significant potential in urban tree and forest inventory by enhancing individual tree extraction \cite{dai2018}, quantifying carbon storage in urban trees \cite{chen2018}, and estimating physiological and biochemical properties such as moisture content and foliar nitrogen \cite{junttila2018, goodbody2020, maltamo2020}. Furthermore, it has been applied to tree species classification and identification of invasive species \cite{yu2017, hakula2023, mielczarek2022}.
The extraction of tree points has gained attention in some studies utilizing MS-LiDAR data.
Chen at al. \cite{chen2018} extracted tree points from MS-LiDAR data by first generating a normalized digital surface model (nDSM) and three intensity images from Optech Titan MS-LiDAR data (1550/1064/532~nm), followed by classifying the data into five land cover classes (house, road, tree, grass, and water) using a traditional support vector machine (SVM) algorithm. Their results showed that incorporating all three spectral channels, pNDVI, and NDWI (normalized difference water index) led to an 11.19 pp increase in overall accuracy (OA) compared to using only mono-spectral (NIR) LiDAR data. However, the raster-based approach and the extraction of tree points among five land cover types are considered key limitations. In a subsequent study, point-based conventional k-means clustering and random forest classification were employed to identify tree points across eight classes (wire, car, road, low vegetation, high vegetation, bare land, housing, and non-residential buildings) before individual tree segmentation \cite{shi2023}. They also investigated the effect of incorporating spectral indices, calculating six of them: difference vegetation index, pNDVI, ratio normalized difference vegetation index, simple ratio, soil-adjusted vegetation index, and enhanced vegetation index 2. Their experiments revealed that while adding individual indices slightly reduced classification accuracy, combining all indices improved OA by 2.9 pp and the kappa coefficient by 2 pp.
More recently, Yang et al. \cite{yang2024} proposed a deep learning-based method for tree point extraction using PTv1 \cite{Zhao2021} architecture, which enables effective learning of both spatial and spectral features from Optech Titan MS-LiDAR data. Their approach demonstrated improvements of 2-5 pp in performance compared to using individual spectral channels. Moreover, Point Transformer outperformed traditional random forest methods and four well-established deep learning models (PointNet, PointNet++, RandLA-Net, and DGCNN). Despite leveraging an advanced deep learning model, the primary limitation of their work is the classification of urban point clouds into six predefined categories (road, grass, building, tree, car, and powerline) rather than focusing on binary tree vs. non-tree segmentation. Additionally, their study did not investigate the impact of spectral indices.


\subsection{Contributions}
Our contributions can be summarized as follows.
\begin{enumerate}
    \item Utilization of MS-LiDAR point clouds for accurate tree point detection in complex urban environments using DL-based binary semantic segmentation.
    \item Presentation of the first publicly available MS-LiDAR urban tree benchmark dataset (Loosdorf-tree).
    \item Incorporation of the point-based pNDVI, derived from laser reflectivity, into the DL framework for enhanced tree detection.
    \item Deployment of the state-of-the-art transformer-based DL models (SPT \cite{robert2023}, PTv1, and PTv3 \cite{wu2024}) for the first time, in the context of binary semantic segmentation of trees.
    \item A rigorous spectral ablation study to evaluate the contribution of active spectral information to tree point detection performance.
\end{enumerate}

\section{METHODOLOGY}

\subsection{Dataset}
The dataset is collected using a multispectral airborne \textit{RIEGL} VQ-1560i-DW laser scanner over the cities of Loosdorf and Melk in Lower Austria, representing both urban and suburban environments. Unlike the Optech Titan, this MS-LiDAR system is currently available on the market and operates at wavelengths of \SI{532}{\nano\meter} (green) and \SI{1064}{\nano\meter} (NIR). The average point densities for the green and NIR scanners are 9.1~points/m$^2$ and 14.7~points/m$^2$, respectively, resulting in an overall multispectral point cloud density of 21.11~points/m$^2$. Detailed specifications of the dataset are provided in Table~\ref{tab:dataset_specs}.
A total of 48,164,790 points ($1.68\ \text{km}^2$) are manually annotated into tree and non-tree classes using CloudCompare software. The dataset contains approximately 18\% tree points, making it imbalanced. The dataset is divided into training, testing, and validation plots in a ratio of 68.53:16.28:15.19. Loosdorf-tree dataset is publicly available via the TU Wien dataset repository\footnote{\url{The_link_to_dataset_will_be_provided_upon_paper_acceptance}}. 
\begin{table*}[ht]
\centering
\caption{Loosdorf-tree dataset specifications.}
\label{tab:dataset_specs}
\begin{tabular}{cccccc}
\hline
\textbf{MS-LiDAR system} & \textbf{Wavelength (nm)} & \textbf{Point density (point/m$^2$)} & \textbf{Pulse repetition rate (kHz)} & \textbf{Beam divergence (mrad)} & \textbf{Flight altitude (m)} \\
\hline
\multirow{2}{*}{VQ-1560i-DW} & 532  & 9.1  & 1000 & 2.19 & \multirow{2}{*}{700} \\
                             & 1064 & 14.69 & 1000 & 0.25 &                        \\
\hline
\end{tabular}
\end{table*}
\subsection{Data Preprocessing}
Individual point clouds are first denoised using the statistical outlier removal (SOR) filter in CloudCompare. They are then merged using the preMergeChannelsPointclouds\footnote{\url{https://opals.geo.tuwien.ac.at/html/stable/preMergeChannelsPointclouds.html}} script in OPALS \cite{pfeifer2014} to create MS-LiDAR point clouds. The reflectance values of the NIR points are derived from seven neighboring green data points within a 1-meter spherical neighborhood, and vice versa for the green points. To perform height normalization, ground points are first identified using the cloth simulation filter (CSF) plugin in CloudCompare. A digital terrain model (DTM) is then generated from these ground points. Normalized height values are calculated by subtracting each point’s elevation from the corresponding DTM height using OPALS. Spectral features are also normalized by applying a combination of outlier-resistant scaling and min–max normalization.
\vspace{-3mm} 
\subsection{Tree Points Extraction}
We utilize range-corrected relative reflectance attributes provided by \textit{RIEGL} for V-Line laser scanners. NIR and green reflectance values in linear units are used for calculating pNDVI, as defined in Equation~\eqref{eq:pndvi} and~\eqref{eq:linear}.
Considering the potential efficiency of transformer-based deep learning models in processing high-dimensional MS-LiDAR data, we investigate two new state-of-the-art transformer architectures for tree point extraction alongside PTv1: SPT and PTv3. SPT and PTv3 DL models have recently been introduced for efficient semantic and instance segmentation of large-scale 3D scenes. SPT first partitions point clouds into a hierarchical superpoint structure using a parallel cut pursuit algorithm, and then learns the semantic relationships between superpoints rather than individual points, making it highly time-efficient. Additionally, it employs a self-attention mechanism to capture relationships between superpoints across multiple scales, thereby enhancing the model's performance. 

PTv3 is the latest version of the Point Transformer models, specifically designed to prioritize simplicity and efficiency over complexity, thus enhancing scalability. Moreover, PTv3 incorporates a number of serialization patterns, such as the Hilbert space-filling curve. By shifting across these serialization patterns, the attention mechanism is able to capture diverse spatial relationships and contextual information, leading to improved model accuracy and generalization. These traits have great potential to significantly enhance the efficiency of high-dimensional MS-LiDAR data processing.

To enable these models to handle MS-LiDAR point clouds containing various spectral information, we modify them by introducing the desired spectral data as new point attributes. We compare the performance of SPT and PTv3 with PTv1, which was identified as the best-performing deep learning model for MS-LiDAR-based land cover classification in \cite{yang2024}. Given the demonstrated superiority of PTv1 over PointNet, PointNet++, RandLA-Net, and DGCNN \cite{yang2024}, we omit comparisons with these four earlier models. The official repositories of SPT\footnote{\url{https://github.com/drprojects/superpoint_transformer}}, PTv3\footnote{\url{https://github.com/Pointcept/PointTransformerV3}} (which also includes PTv1) are used to implement them.

 To ensure a fair comparison, all models are trained for 300 epochs and use AdamW optimizer, which is the default choice for many transformer-based DL models and offers better generalizability than the standard Adam optimizer and weighted cross-entropy loss, adjusted according to class frequency. The size of the grid is fixed at \SI{0.1}{\meter}. The input features include green and NIR reflectances in addition to coordinates. Detailed descriptions of the main hyperparameters are provided in the following subsections. 

\subsubsection{SPT}
We include reflectance values from the green and NIR channels alongside geometric information during both the superpoint generation and training steps. Empirically, the number of nearest neighbors is fixed at 30, with a search radius of \SI{10}{\meter}. In SPT, the most important parameters are regularization, spatial weight, and cutoff, which are the inputs of the cut pursuit partitioning algorithm in order to create superpoints. The values of regularization, spatial weight, and cutoff are empirically set to [0.1, 0.2, 0.2], [0.1, 0.01, 0.001], and [10, 20, 60], respectively. The learning rate is set to 0.01, and the weight decay is 0.0001.

\subsubsection{PTv1}
The PTv1 hyperparameters are configured following the settings proposed by Yang et al.~\cite{yang2024}, with a batch size of 6, a maximum of 65,536 points per batch, a learning rate of 0.001, and a weight decay of 0.003. The PointTransformer-Seg50 architecture is employed.

\subsubsection{PTv3}
Four serialization patterns—Z-order, Trans Z-order, Hilbert curves, and Trans Hilbert—are applied within the model architecture with a multilayer perceptron ratio of 6 and 64 output channels. The common parameters of PTv3 are configured similarly to PTv1.

\begin{equation}
\text{pNDVI} = \frac{\text{NIR}_{\text{linear}} - \text{green}_{\text{linear}}}{\text{NIR}_{\text{linear}} + \text{green}_{\text{linear}}}
\label{eq:pndvi}
\end{equation}
\begin{equation}
\text{Reflectance}_{\text{linear}} = 10^{\frac{\text{Reflectance}_{\text{dB}}}{10}}
\label{eq:linear}
\end{equation}

\section{EXPERIMENTAL RESULTS}
The results are validated using established metrics, including OA, mean accuracy (mAcc), IoU, and mIoU. All experiments are conducted on a Linux Mint 21.3 system equipped with an NVIDIA L40 GPU featuring 48.31~GB of onboard memory and a 32-core AMD EPYC 9354 CPU.

Table~\ref{tab:DLmodels_validation} compares the performance of transformer-based DL models. According to the results, SPT, PTv3, and PTv1 achieve the highest accuracies for tree detection, respectively. Although all transformer-based models obtain high accuracy (mIoU $>$ 81.96\%), SPT yields a tree IoU exceeding those of PTv1 and PTv3 by 4.5 pp and 2.5 pp, respectively.
The code and pre-trained model of the best-performing tree extraction DL model (SPT) for both mono- and MS point clouds are publicly available in the paper’s GitHub repository \footnote{\url{https://github.com/3DOM-FBK/TreePointExtractor}} to facilitate further usage and development. 
\begin{table}[h!]
    \centering
    \caption{Comparison of Tree Points Extraction Results from MS-LiDAR (XYZ + Green + NIR) Using Different Deep Learning Methods.}
    \label{tab:DLmodels_validation}
    \begin{tabular}{lccc}
        \hline
        & SPT & PTv1 & PTv3 \\ \hline
        \(\text{IoU}_{\text{Non-tree}}\)~(\%) & \textbf{93.03} & 90.88 & 91.85 \\\hline
        \(\text{IoU}_{\text{Tree}}\)~(\%) & \textbf{77.54} & 73.04 & 75.04 \\\hline
        mIoU (\%) & \textbf{85.28} & 81.96 & 83.44\\\hline
        mAcc (\%) & \textbf{92.14} & 91.74 & 92.02 \\\hline
        OA (\%) &  \textbf{94.38} & 92.69 &  93.45\\\hline
    \end{tabular}
\end{table}

\section{DISCUSSION}
\subsection{Spectral Ablation Study}

\begin{figure*}[h]
    \centering
    \includegraphics[width=0.85
    \textwidth]{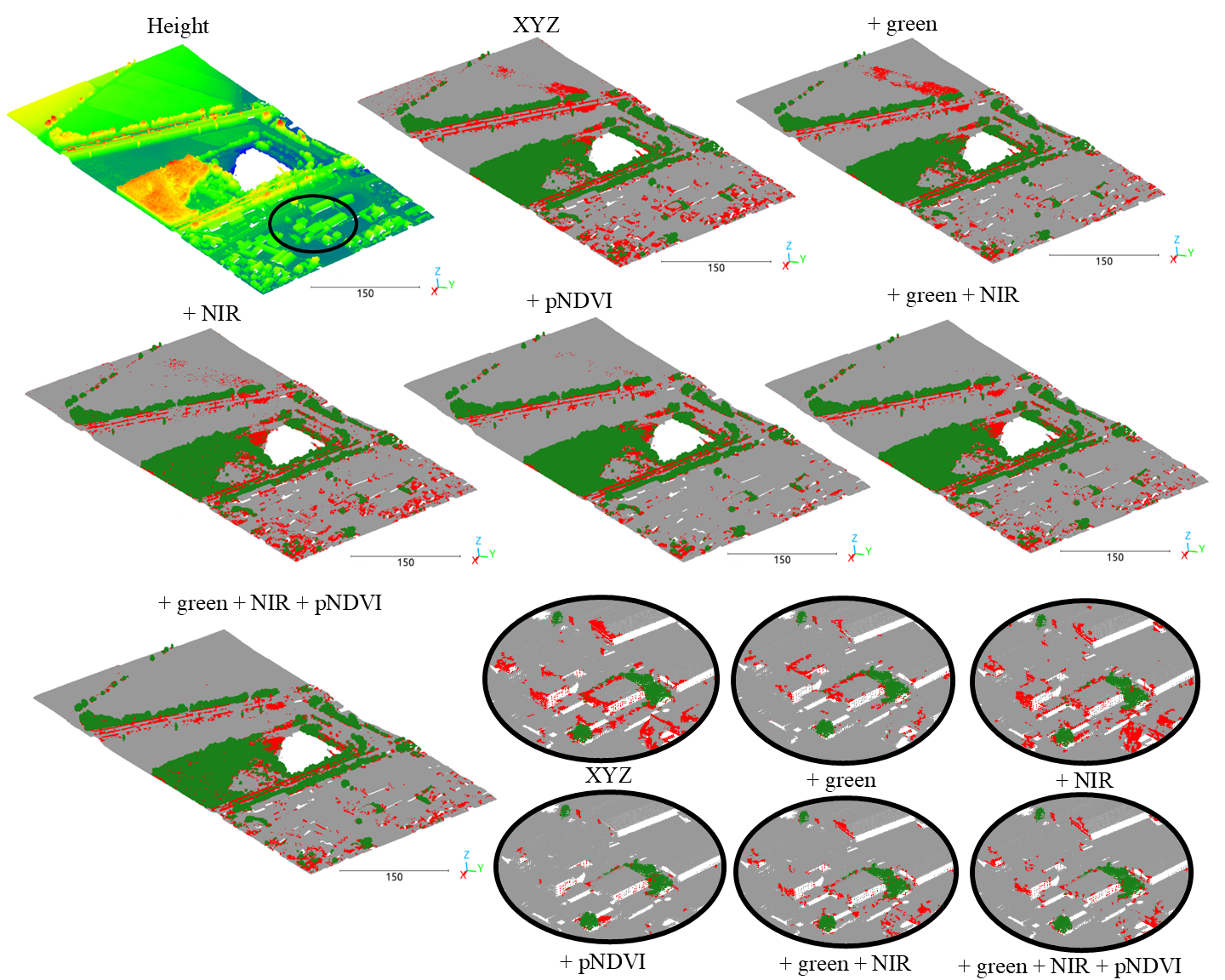}
    \caption{Spectral ablation study on tree point extraction using the superior DL model (SPT): a comparison of using only spatial information, spatial-mono-spectral, and spatial-MS information. Tree extraction errors are highlighted in red, while correctly extracted trees are shown in green.}
    \label{fig:ablation_study}
\end{figure*}

To comprehensively examine the effect of laser spectral information, we conduct a spectral ablation study by evaluating six initial feature vectors used as the input of DL models. Unlike \cite{yang2024}, we further substantiate the advantages of MS-LiDAR data by analyzing the impact of the pNDVI vegetation index. Table~\ref{tab:ablation_study} presents the results of the spectral ablation study based on the best-performing DL model (SPT), and Figure~\ref{fig:ablation_study} visualizes the effect of adding spectral information on tree points extraction. 

In tree detection, while certain errors — such as those caused by low and medium vegetation — can be easily resolved through height thresholding (usually below \SI{2}{\meter}), misclassified objects with greater heights, particularly those with geometries similar to trees, such as facades, cables, electric towers, and fences, pose the most serious challenge. These non-tree classes cannot be easily eliminated through post-processing. Therefore, to precisely evaluate the effect of spectral information on tree point extraction for practical usage, we calculate the error rate of points with normalized heights above \SI{2}{\meter}, as shown in Table~\ref{tab:ablation_study_greater2}.

According to Table~\ref{tab:ablation_study}, SPT effectively detects trees across all spatial and spatial–spectral configurations, achieving mIoU scores exceeding 70.29\%. The combination of coordinates, green, and NIR yields the highest mIoU (85.28\%) and mAcc (92.14\%). Incorporating three spectral features (i.e., green, NIR, and pNDVI) results in the next highest mIoU with a slight difference (84.95\%). Among single spectral feature combinations, pNDVI achieves the highest mIoU (84.45\%). On the other hand, while all spectral features, including the monochromatic green wavelength, improve tree point extraction by more than 2 pp, incorporating only the NIR wavelength is not effective for detecting tree points in complex urban areas, reducing IoU by 1.5 pp compared to using coordinates alone.

Having said that, as mentioned previously, the error rate associated with points having a normalized height above \SI{2}{\meter} is more important for tree detection than general metrics. Both Figure~\ref{fig:ablation_study} and Table~\ref{tab:ablation_study_greater2} demonstrate the substantial superiority of using pNDVI alone, reducing the error rate by 10.61 pp. Since pNDVI is derived from NIR and green wavelengths, it shows the promising potential of MS-LiDAR data in enhancing tree point extraction. Furthermore, using a spectral feature derived from two wavelengths, rather than directly incorporating individual wavelengths, accelerates MS-LiDAR data processing.

\begin{table}[h]
    \centering
    \small 
    \setlength{\tabcolsep}{3pt} 
    \renewcommand{\arraystretch}{0.9} 
    \caption{Effect of Spectral Information of MS-LiDAR on Tree Point Extraction based on the best-performing DL model (SPT).}
    \label{tab:ablation_study}
    \begin{tabular}{lccccc}
        \hline
        & \multicolumn{5}{c}{(\%)} \\[-1pt] 
        \cline{2-6}
        & \shortstack{IoU\\Non-tree} 
        & \shortstack{IoU\\Tree} 
        & mIoU 
        & mAcc 
        & OA \\ 
        \hline
        XYZ & 90.66 & 72.76 & 81.71 & 91.88 & 92.53\\ 
        +~green & 92.13 & 75.45 & 83.79 & 91.79 & 93.66\\ 
        +~NIR & 90.13 & 70.29 & 80.21 & 89.84 & 92 \\ 
        +~pNDVI & 92.76 & 76.15 & 84.45 & 90.96 & 94.12\\       
        +~green +~NIR & \textbf{93.03} & \textbf{77.54} & \textbf{85.28} & \textbf{92.14} & \textbf{94.38}\\ 
        +~green +~NIR +~pNDVI & 93 & 76.90 & 84.95 & 91.27 & 94.33 \\
        \hline
    \end{tabular}
\end{table}

\begin{table}[h]
\centering
\caption{Tree extraction error rates of points with normalized height above \SI{2}{\meter} using SPT.}
\label{tab:ablation_study_greater2}
\begin{tabular}{lccccc}
\hline
XYZ & +~green & +~NIR & +~pNDVI & \shortstack{+~green\\+~NIR} & \shortstack{+~green\\+~NIR\\+~pNDVI} \\
\hline
39.63\% & 31.07\% & 41.57\% & \textbf{29.02\%} & 35.12\% & 35.93\% \\   
\hline
\end{tabular}
\end{table}

\subsection{Time Efficiency}
Processing multispectral point clouds is inherently time-consuming. Table~\ref{tab:TimeEfficiency} presents a comparison of the computational efficiency of the studied DL models when initialized with the incorporation of basic MS-information (NIR and green reflectances) and trained on a single GPU. Among the transformer-based models, PTv1 not only exhibits the lowest accuracy but also has the highest number of training parameters and is by far the most time-consuming. In contrast, SPT is the most time-efficient model with the fewest trainable parameters. Consequently, SPT has great potential for time-efficient and accurate processing of MS-LiDAR data.
\begin{table}[h]
    \centering
    \caption{Efficiency Comparison Among Deep Learning Models.}
    \label{tab:TimeEfficiency}
    \begin{tabular}{cccc}
        \hline
        & SPT & PTv1 & PTv3 \\ \hline
        Parameters & 212K & 170.3M & 46.15M \\ \hline
        Elapsed time (h) & 0.67 & 26.37 & 2.32 \\ \hline
    \end{tabular}
\end{table}

\section{Conclusions}
In this paper, we propose a single-data-source and class definition-agnostic approach for accurately extracting tree areas in complex urban and suburban environments by synergistically leveraging multispectral LiDAR data and deep learning-based binary semantic segmentation. Since trees are a vital component of urban green infrastructure, their automatic mapping is becoming increasingly important for effective city management and planning. Automatically extracting tree points is a crucial prior step in accurate urban tree inventories. Due to the presence of diverse objects across different geographic regions and the resulting inconsistencies in class definitions, class-specific deep learning models are gaining increasing attention.

Our experiments substantiate the notable merit of leveraging laser spectral information. Notably, even single-wavelength information of conventional LiDAR boosts tree points extraction. It should be outlined that although active multispectral information demonstrates improved tree point detection in complex environments, it provides even greater benefits for urban tree management, particularly for tree species classification and health monitoring.

The results obtained demonstrate the promising potential of the relative reflectance values provided by \textit{RIEGL} V-Line scanners. However, since these values do not consider the incidence angle, improved outcomes could be achieved by incidence angle correction. Introduced as the first commercial multispectral airborne LiDAR system, Optech Titan collects point cloud data at three distinct wavelengths and has been widely adopted in multispectral laser scanning research. Nonetheless, Optech Titan lacks the ability to deliver relative reflectivity and is now out of production. Thus, attention must shift toward existing MS-LiDAR systems to assess their functionalities. Further improvements can be achieved by deploying multispectral LiDAR systems that operate across a greater number of channels. Additionally, our experiments reveal the importance of developing time-efficient and, at the same time, accurate deep learning models like SPT for processing voluminous MS point clouds.

\section*{Acknowledgment}
This research is funded by the European Spatial Data Research (EuroSDR). We also gratefully acknowledge the support of \textit{RIEGL} in the data acquisition and research Council of Finland project 359203.
\ifCLASSOPTIONcaptionsoff
  \newpage
\fi

\bibliographystyle{IEEEtran}
\bibliography{library}

\begin{thebibliography}{10}
\providecommand{\url}[1]{#1}
\csname url@samestyle\endcsname
\providecommand{\newblock}{\relax}
\providecommand{\bibinfo}[2]{#2}
\providecommand{\BIBentrySTDinterwordspacing}{\spaceskip=0pt\relax}
\providecommand{\BIBentryALTinterwordstretchfactor}{4}
\providecommand{\BIBentryALTinterwordspacing}{\spaceskip=\fontdimen2\font plus
\BIBentryALTinterwordstretchfactor\fontdimen3\font minus \fontdimen4\font\relax}
\providecommand{\BIBforeignlanguage}[2]{{%
\expandafter\ifx\csname l@#1\endcsname\relax
\typeout{** WARNING: IEEEtran.bst: No hyphenation pattern has been}%
\typeout{** loaded for the language `#1'. Using the pattern for}%
\typeout{** the default language instead.}%
\else
\language=\csname l@#1\endcsname
\fi
#2}}
\providecommand{\BIBdecl}{\relax}
\BIBdecl

\bibitem{yang2024}
J.~Yang, R.~Gan, B.~Luo, A.~Wang, S.~Shi, and L.~Du, ``An improved method for individual tree segmentation in complex urban scene based on using multispectral lidar by deep learning,'' \emph{IEEE Journal of Selected Topics in Applied Earth Observations and Remote Sensing}, pp. 1--17, 2024.

\bibitem{Hyypaa2013}
A.~Hyypp{\"a}, Y.~Jaakkola, A.~Chen, A.~Kukko, and H.~Kaartinen, ``Unconventional lidar mapping from air, terrestrial and mobile,'' in \emph{proc. Photogrammetric Week}.\hskip 1em plus 0.5em minus 0.4em\relax Berlin, Germany: Wichmann/VDE Verlag, 2013, pp. 205--214.

\bibitem{takhtkeshha2024}
N.~Takhtkeshha, G.~Mandlburger, F.~Remondino, and J.~Hyyppä, ``Multispectral light detection and ranging technology and applications: A review,'' \emph{Sensors}, vol.~24, no.~5, p. 1669, 2024.

\bibitem{hakula2023}
A.~Hakula, L.~Ruoppa, M.~Lehtomäki, X.~Yu, A.~Kukko, H.~Kaartinen, J.~Taher, L.~Matikainen, E.~Hyyppä, V.~Luoma, M.~Holopainen, V.~Kankare, and J.~Hyyppä, ``Individual tree segmentation and species classification using high-density close-range multispectral laser scanning data,'' \emph{ISPRS Open Journal of Photogrammetry and Remote Sensing}, vol.~9, p. 100039, 2023.

\bibitem{RUOPPA2025}
L.~Ruoppa, O.~Oinonen, J.~Taher, M.~Lehtomäki, N.~Takhtkeshha, A.~Kukko, H.~Kaartinen, and J.~Hyyppä, ``Unsupervised deep learning for semantic segmentation of multispectral lidar forest point clouds,'' \emph{ISPRS Journal of Photogrammetry and Remote Sensing}, vol. 228, pp. 694--722, 2025.

\bibitem{adityaBenchmark2024}
A.~Aditya, B.~Lohani, J.~Aryal, and S.~Winter, ``Benchmarking deep learning architectures for urban vegetation point cloud semantic segmentation from mls,'' \emph{IEEE Trans. Geoscience and Remote Sensing}, vol.~62, pp. 1--1, 2024.

\bibitem{adityaGreenSegNet2024}
------, ``Greensegnet: A novel deep learning architecture for urban vegetation segmentation from mls data,'' \emph{IEEE Transactions on Geoscience and Remote Sensing}, vol.~62, pp. 1--10, 2024.

\bibitem{dai2018}
W.~Dai, B.~Yang, Z.~Dong, and A.~Shaker, ``A new method for 3d individual tree extraction using multispectral airborne lidar point clouds,'' \emph{ISPRS Journal of Photogrammetry and Remote Sensing}, vol. 144, 2018.

\bibitem{chen2018}
X.~Chen, C.~YE, J.~Li, and M.~Chapman, ``Quantifying the carbon storage in urban trees using multispectral als data,'' \emph{IEEE Journal of Selected Topics in Applied Earth Observations and Remote Sensing}, pp. 1--8, 2018.

\bibitem{junttila2018}
S.~Junttila, J.~Sugano, M.~Vastaranta, R.~Linnakoski, H.~Kaartinen, A.~Kukko, M.~Holopainen, H.~Hyyppä, and J.~Hyyppä, ``Can leaf water content be estimated using multispectral terrestrial laser scanning? a case study with norway spruce seedlings,'' \emph{Frontiers in Plant Science}, vol.~9, p. 299, 2018.

\bibitem{goodbody2020}
T.~Goodbody, P.~Tompalski, N.~Coops, C.~Hopkinson, P.~Treitz, and K.~van Ewijk, ``Forest inventory and diversity attribute modelling using structural and intensity metrics from multi-spectral airborne laser scanning data,'' \emph{Remote Sensing}, vol.~12, p. 2109, 2020.

\bibitem{maltamo2020}
M.~Maltamo, J.~Räty, L.~Korhonen, E.~Kotivuori, M.~Kukkonen, H.~Peltola, J.~Kangas, and P.~Packalen, ``Prediction of forest canopy fuel parameters in managed boreal forests using multispectral and unispectral airborne laser scanning data and aerial images,'' \emph{European Journal of Remote Sensing}, vol.~53, p. 245–257, 2025.

\bibitem{yu2017}
X.~Yu, J.~Hyyppä, P.~Litkey, H.~Kaartinen, M.~Vastaranta, and M.~Holopainen, ``Single-sensor solution to tree species classification using multispectral airborne laser scanning,'' \emph{Remote Sensing}, vol.~9, p. 108, 2017.

\bibitem{mielczarek2022}
D.~Mielczarek, P.~Sikorski, P.~Archiciński, W.~Ciężkowski, E.~Zaniewska, and J.~Chormanski, ``The use of an airborne laser scanner for rapid identification of invasive tree species acer negundo in riparian forests,'' \emph{Remote Sensing}, vol.~15, p. 212, 2022.

\bibitem{shi2023}
S.~Shi, X.~Tang, B.~Chen, B.~Chen, Q.~Xu, S.~Bi, and W.~Gong, ``Point cloud data processing optimization in spectral and spatial dimensions based on multispectral lidar for urban single-wood extraction,'' \emph{ISPRS International Journal of Geo-Information}, vol.~12, p.~90, 2023.

\bibitem{Zhao2021}
H.~Zhao, L.~Jiang, J.~Jia, P.~Torr, and V.~Koltun, ``Point transformer,'' in \emph{Proc. ICCV}, 2021, pp. 16\,239--16\,248.

\bibitem{robert2023}
D.~Robert, H.~Raguet, and L.~Landrieu, ``Efficient 3d semantic segmentation with superpoint transformer,'' in \emph{Proc. ICCV}, 2023, pp. 1785--1795.

\bibitem{wu2024}
X.~Wu, L.~Jiang, P.-S. Wang, Z.~Liu, X.~Liu, Y.~Qiao, W.~Ouyang, H.~Tong, and H.~Zhao, ``Point transformer v3: Simpler, faster, stronger,'' in \emph{Proc. CVPR}, 2024, pp. 4840--4851.

\bibitem{pfeifer2014}
N.~Pfeifer, G.~Mandlburger, J.~Otepka, and W.~Karel, ``Opals—a framework for airborne laser scanning data analysis,'' \emph{Computers, Environment and Urban Systems}, vol.~45, p. 125–136, 2014.

\end{thebibliography}

\end{document}